\documentclass[]{fairmeta}
\usepackage{microtype}
\usepackage{amsfonts}
\usepackage{graphicx}
\usepackage{booktabs}
\usepackage{wrapfig}
\usepackage{amsmath}
\usepackage{amsthm}
\usepackage{algpseudocode}
\usepackage[linesnumbered,lined,boxed,commentsnumbered,ruled,longend]{algorithm2e}
\usepackage[capitalize,noabbrev]{cleveref}
\theoremstyle{plain}
\newtheorem{theorem}{Theorem}[section]

\theoremstyle{definition}

\theoremstyle{remark}

\usepackage{enumitem}
\usepackage[subtle, mathdisplays=tight, charwidths=tight, leading=normal]{savetrees}
\usepackage{subcaption} % Recommended alternative
\usepackage{booktabs} % for professional tables
\usepackage{longtable}
\usepackage{textcomp}

\SetKwInput{KwInput}{Input}
\SetKwInput{KwOutput}{Output}

\usepackage{enumitem}
\newenvironment{denseitemize}{
\begin{itemize}[topsep=0.5pt, partopsep=0pt, leftmargin=2.5em]
  \setlength{\itemsep}{2.5pt}
  \setlength{\parskip}{0pt}
  \setlength{\parsep}{0pt}
}{\end{itemize}}

\usepackage{tcolorbox}
\usepackage{xcolor}

\tcbset{
  myexample/.style={
    colback=white,
    colframe=black!60,
    colbacktitle=black!80,
    coltitle=white,
    fonttitle=\bfseries,
    arc=4mm,
    boxrule=1pt,
    left=3mm, right=3mm, top=1.5mm, bottom=1.5mm,
    title={#1}
  }
}

\definecolor{caseblue}{HTML}{5C6FA8}

\tcbset{
  casestudy/.style={
    colback=white,
    colframe=black!60,
    colbacktitle=caseblue,    % set background color of title
    coltitle=white,
    fonttitle=\bfseries,
    arc=4mm,
    boxrule=1pt,
    left=3mm, right=3mm, top=1.5mm, bottom=1.5mm,
    title={#1}
  }
}
\usepackage{multirow}
\usepackage{makecell}

\usepackage{tikz} % Include in your preamble if not already there

\title{
{\fontsize{16pt}{10pt}\selectfont
    Prosperity before Collapse: How Far Can Off-Policy RL Reach with Stale Data on LLMs?
}
}

\definecolor{skyblue}{RGB}{135, 206, 235}
\definecolor{palegreen}{RGB}{152, 251, 152}

\author{
Haizhong Zheng$^{1}$,
Jiawei Zhao$^{2}$,
Beidi Chen$^{1}$ \\
$^1$Carnegie Mellon University \\
$^2$Meta AI \\
\texttt{\{haizhonz, beidic\}@cmu.edu},
\texttt{jwzhao@meta.com} \\
}

\abstract{
Reinforcement learning has been central to recent advances in large language model reasoning, but most algorithms rely on on-policy training that demands fresh rollouts at every update, limiting efficiency and scalability. Asynchronous RL systems alleviate this by decoupling rollout generation from training, yet their effectiveness hinges on tolerating large staleness in rollout data, a setting where existing methods either degrade in performance or collapse. We revisit this challenge and uncover a \emph{prosperity-before-collapse} phenomenon: stale data can be as informative as on-policy data if exploited properly. Building on this insight, we introduce \textbf{M2PO} (Second-Moment Trust Policy Optimization), which constrains the second moment of importance weights to suppress only extreme outliers while preserving informative updates. Notably, M2PO sharply reduces the fraction of clipped tokens under high staleness (from 1.22\% to 0.06\% over training), precisely masking high-variance tokens while maintaining stable optimization. Extensive evaluation across six models (from 1.7B to 32B) and eight benchmarks shows that M2PO delivers stable off-policy training even with data stale by \underline{\emph{at least 256 model updates}} and matches on-policy performance.
}

\metadata[Github]{\url{https://github.com/Infini-AI-Lab/M2PO/}}
\metadata[Website]{\url{https://infini-ai-lab.github.io/M2PO/}}

\begin{document}

\maketitle

\vspace{0.45cm}
\begin{figure}[h]
  \centering
  \includegraphics[width=0.98\linewidth]{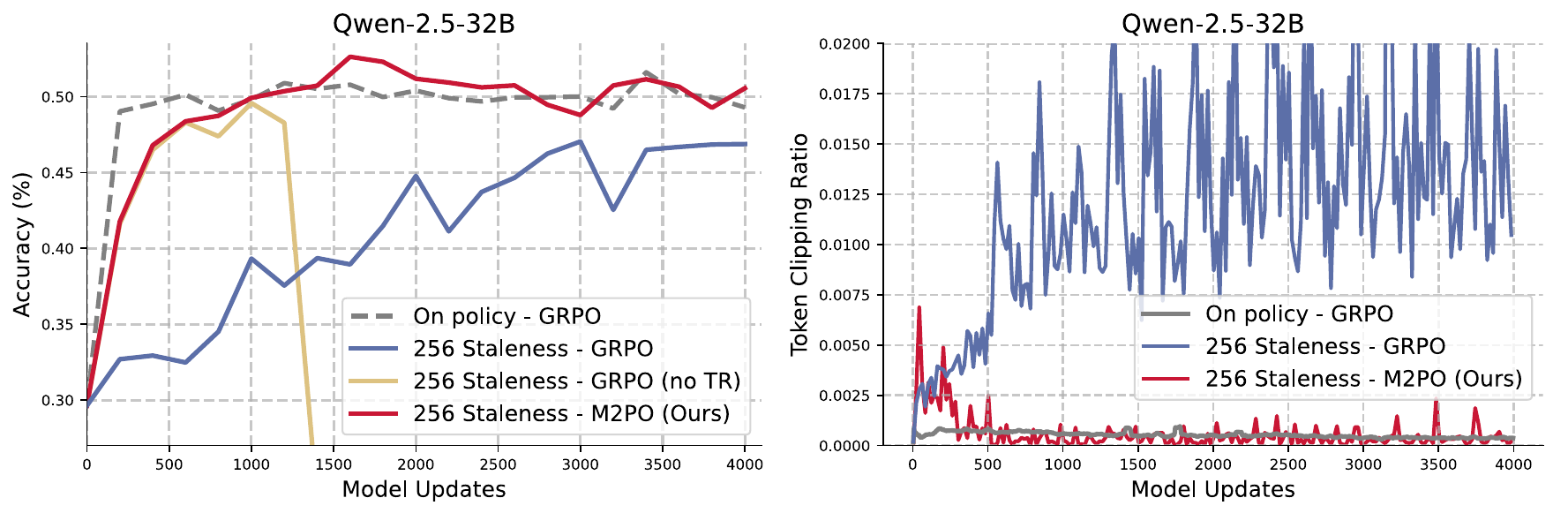} 
\caption{Comparison of on-policy GRPO and off-policy training under a staleness of 256 model updates on Qwen-2.5-32B. 
\textbf{Left:} Standard GRPO suffers from degradation with stale rollouts, while removing the trust region (GRPO no TR) reveals a clear \textit{prosperity-before-collapse} phenomenon. 
In contrast, M2PO achieves stable training and matches on-policy performance even under high staleness. 
\textbf{Right:} Token clipping ratio comparison shows that M2PO dramatically reduces clipping events compared to GRPO with the same staleness, while avoiding training collapse.}
  \label{fig:fig1}
\end{figure}

\section{Introduction}
% \vspace{-0.2cm}
Reinforcement learning (RL) has been central to recent advances in large language model (LLM) reasoning, driving breakthroughs in systems like OpenAI’s o1~\citep{openai2024openaio1card} and DeepSeek’s R1~\citep{deepseekai2025deepseekr1incentivizingreasoningcapability, team2025kimi}.
Most existing RL algorithms~\citep{schulman2017proximal, zheng2025act, yu2025dapo} for LLMs adopt an on-policy design, as it provides stable training and reliable performance, but the strict requirement for fresh~(or limited-staleness) rollouts at every update leads to substantial inefficiency and limits scalability.
To overcome this bottleneck, a growing line of RL systems~\citep{fu2025areal, slime_github, noukhovitch2024asynchronous, zhong2025streamrl, he2025history} have explored asynchronous designs that decouple rollout from training.
Such approaches improve resource utilization and enable training to scale more efficiently across large and heterogeneous clusters, but their effectiveness fundamentally relies on the ability of RL algorithms to tolerate rollout staleness without sacrificing stability or performance.

However, under large rollout staleness, existing RL algorithms struggle to strike the right balance. Some methods~\citep{schulman2017proximal, shao2024deepseekmath, zheng2025group} can maintain stability, but they often suffer from noticeable performance degradation. Conversely, approaches designed to maximize performance~\citep{fu2025areal, chen2025minimax, su2025klear} tend to compromise stability, frequently leading to training collapse. On-policy methods provide both stability and strong performance, but their reliance on fresh or only slightly stale rollouts at every update imposes rigid constraints that hinder scalability.
Consequently, an ideal off-policy RL algorithm for LLMs should enable effective reuse of trajectories collected under outdated policies to preserve strong performance under significant staleness, and ensure stable training that converges competitively with on-policy methods. 
Meeting these requirements is key to realizing off-policy RL as a truly scalable solution for aligning and fine-tuning large language models. 

In this paper, we aim to investigate the underlying reasons for the limitations of off-policy RL in LLMs and to design an effective algorithm that fully leverages stale data to unlock its potential.
We begin by revealing an intriguing \emph{Prosperity before Collapse} phenomenon~(Yellow curve in Figure~\ref{fig:fig1} (left)): although RL training without a trust region eventually collapses on stale data, it initially achieves substantially higher performance than vanilla GRPO with $\epsilon$-clipping. In some cases, it even matches the performance of the on-policy baseline.
From this, we draw an important observation: \emph{stale data can be as informative as data collected on-policy in RL for LLMs}, but the key challenge lies in how existing algorithms exploit it. In particular, vanilla GRPO performs poorly under staleness because stale-data training exhibits a substantially higher clipping rate, with many of the clipped updates occurring on informative high-entropy tokens~(see Figure~\ref{fig:token-masking}).
This disproportionate clipping on crucial tokens hinders the full utilization of stale training data.

This pivotal token masking observation reveals that these high-entropy tokens play a dual role: they provide the most informative training signal but also introduce the greatest instability under staleness. 
Therefore, the key challenge is to retain as much learning signal from these tokens as possible without risking training collapse.
Motivated by this, we propose \textbf{M2PO} (Second-Moment Trust Policy Optimization), a novel off-policy RL algorithm that constrains the second moment of importance weights. 
Unlike standard $\epsilon$-clipping, which disproportionately suppresses high-entropy tokens and discards valuable learning signals, M2PO leverages the second-moment metric $M_2$. 
This metric is both variance-sensitive, capturing instability introduced by high-entropy tokens, and statistically stable, avoiding the cancellation issues inherent to KL-based measures. 
By regularizing training at the batch level through $M_2$, M2PO masks only extreme outliers while preserving the majority of informative updates.
As a result, M2PO enables stable off-policy reinforcement learning with stale data, matching on-policy performance even under large staleness.

% As illustrated by Figure~\ref{fig:fig1}, highlighted by Figure~\ref{fig:fig1}~(left), even only trained with data stale by at least 256 model updates, M2PO achieves accuracy comparable to the on-policy baseline across all models~(red curve), highlighting its ability to fully exploit stale data without sacrificing stability.
% M2PO achieves this by applying a more accurate and adaptive clipping strategy to clip much fewer tokens while maintaining the training stability.
% As shown in Figure~\ref{fig:fig1} (right), M2PO dramatically reduces the fraction of clipped tokens under high staleness (from 1.22\% to 0.06\%), 
% To better verify the effectiveness of M2PO, we conduct an extensive evaluation of M2PO across six model scales~(from 1.7B to 32B) and eight math reasoning benchmarks. 
% Our evaluation results show that M2PO consistently achieves strong performance across all training settings.
% Finally, another favorable property of M2PO is its insensitivity to the choice of threshold: a single value is used for all experiments and consistently delivers strong results, showing its practicality and robustness.

As illustrated in Figure~\ref{fig:fig1}~(left), even when trained exclusively on data stale by at least 256 model updates, M2PO achieves accuracy comparable to the on-policy baseline~(red curve), demonstrating its ability to fully exploit stale data without sacrificing stability.
M2PO achieves this through a more accurate and adaptive clipping strategy that clips substantially fewer tokens while maintaining training stability.
As shown in Figure~\ref{fig:fig1}~(right), M2PO dramatically reduces the fraction of clipped tokens under high staleness~(from 1.22\% to 0.06\% over the entire training process, see Figure~\ref{fig:clipping-bar}), thereby preserving more useful training information in stale data.
To further validate M2PO effectiveness, we conduct an extensive evaluation of M2PO across six model scales (ranging from 1.7B to 32B) and eight math reasoning benchmarks in Section~\ref{sec:exp}.
The results show that M2PO consistently delivers strong performance across all training settings.
M2PO also shows insensitivity to the choice of threshold, with a single value across all experiments, demonstrating its practicality and robustness.

% In Section~\ref{sec:exp}, we present an extensive evaluation of M2PO across six model scales~(from 1.7B to 32B) and eight math reasoning benchmarks. 
% First, our experiments demonstrate that M2PO remains effective even under extreme staleness, where standard methods typically collapse or achieve inferior performance. 
% In particular, even only trained with data stale by at least 256 model updates, M2PO achieves accuracy comparable to the on-policy baseline across all models~~(Red curve in Figure~\ref{fig:fig1} (left)), highlighting its ability to fully exploit stale data without sacrificing stability. 

% \section{Related Work}
\section{Related Work}

\textbf{RLVR.}
% \textbf{RL for LLM Reasoning.} 
% Reinforcement learning~(RL) was initially used to align model outputs with human preferences~\citep{ouyang2022training, dai2023safe}.
% Since then, RL has become a commonly used technique for fine-tuning LLMs, enabling them to generate more helpful, harmless, and honest responses by incorporating reward signals from human feedback~\citep{christiano2017deep, bai2022training}. 
Recent advances~\citep{deepseekai2025deepseekr1incentivizingreasoningcapability, yu2025dapo, team2025kimi, gao2024designing} in LLM reasoning show that Reinforcement Learning with Verifiable Reward~(RLVR), which relies on verifiable reward signals instead of model-generated scores, can effectively improve model reasoning ability.
These gains are achieved using various policy optimization methods such as PPO~\citep{ouyang2022training} and GRPO~\citep{shao2024deepseekmath}.
Encouraged by the success of RLVR, a growing body of work~\citep{kazemnejad2024vineppo, yuan2025vc_ppo, yuan2025vapo, yu2025dapo, liu2025understanding, deepcoder2025, zhang2025srpo, hu2025reinforce_pp, xiong2025minimalist} has emerged to further improve reinforcement learning methods for LLM reasoning.
For instance, methods such as VinePPO~\citep{kazemnejad2024vineppo}, VC-PPO~\citep{yuan2025vc_ppo}, and VAPO~\citep{yuan2025vapo} aim to enhance LLM reasoning by optimizing the value function.
% Meanwhile, DAPO~\citep{yu2025dapo} introduces several techniques to improve GRPO, including Dynamic Sampling, which filters out zero-variance prompts and refills the training batch with effective training data through resampling.
% \hz{this is the related work from greso -- add new papers in}

\textbf{Trust Region in RLVR.}
While RLVR has been widely adopted for fine-tuning LLMs, a key challenge lies in how to effectively constrain the trust region, not only to stabilize training but also to achieve better learning efficiency and overall performance.
To address this, a growing line of work has proposed various strategies to control the policy update, ranging from ratio clipping~\citep{yu2025dapo}, approximate trust region~\citep{fu2025areal}, sequence-level clipping~\citep{zheng2025group}, asymmetric trust region~\citep{roux2025tapered, arnal2025asymmetric}, and gradient-preserving clipping~\citep{su2025klear, chen2025minimax}.
For instance,
AREAL~\citep{fu2025areal} uses a more recent approximate policy to decide the trust region rather than the behavior model.
GSPO~\citep{zheng2025group} moves from token-level to sequence-level clipping by defining importance ratios on sequence likelihood.
While these methods improve RLVR under moderate settings, most of them focus on relatively limited intra-iteration staleness~(e.g., 8 or 16) and have not been thoroughly studied under larger off-policy gaps, like extreme staleness.
In this work, our goal is to better understand the role of staleness in RLVR and 
to seek more effective ways of constraining the trust region in RLVR.

\vspace{-0.1cm}
\section{Background}

\subsection{Group Relative Policy Optimization (GRPO)}

Group Relative Policy Optimization~(GRPO)~\citep{shao2024deepseekmath} is a variant of Proximal Policy Optimization~(PPO)~\citep{ouyang2022training} tailored for language model fine-tuning. Instead of computing advantages using a value function, GRPO normalizes reward scores within groups of responses sampled for the same prompt, which largely improves the training efficiency, and aims to maximize the following objective:
% \vspace{-0.5cm}

\begin{equation}
\begin{split}
    \mathcal{J}_{GRPO}&(\theta) = \mathbb{E}{[q \sim P(Q), \{o_i\}_{i=1}^G \sim \pi_{\theta_{old}}(O|q)]}  \\
    & \frac{1}{G}\sum_{i=1}^G \left( \min \left( \frac{\pi_\theta(o_i |q)}{\pi_{\theta_{behav}}(o_i |q)} A_i, \text{clip} \left( \frac{\pi_\theta(o_i |q)}{\pi_{\theta_{behav}}(o_i |q)}, 1 - \epsilon, 1 + \epsilon \right)  A_i \right) \right) ,
\end{split}
\label{eq:GRPO-obj}
\end{equation}

% where $A_i$ is the advantage, computed using a group of rewards $\{r_1, r_2, \ldots, r_G\}$ corresponding to the outputs within each group:
where $A_i$ is the advantage, computed using a group of rewards corresponding to the outputs within each group:
% \vspace{-0.4cm}

\begin{equation}
\label{eq:group-adv}
A_{i,t} = \frac{r_i - \text{mean}(\{R_i\}_{i=1}^G)}{\text{std}(\{R_i\}_{i=1}^G)}. 
\end{equation} 
% \vspace{-0.2cm}

Similar to PPO, GRPO employs a clipping mechanism to stabilize updates. 
The ratio $r_i=\tfrac{\pi_\theta(o_i|q)}{\pi_{\theta_{old}}(o_i|q)}$ is clipped to $[1-\epsilon, 1+\epsilon]$, 
so that when $A_i>0$ the policy cannot increase probability mass excessively, 
and when $A_i<0$ it cannot over-penalize. 
This prevents large, unstable updates while still allowing normalized group advantages to guide learning.

% \vspace{-0.2cm}
\subsection{Performance Degradation from Training with Stale Data}
\label{ssec:stale-performance}
% Introduce PPO and GRPO.

\begin{wrapfigure}{r}{0.31\textwidth}
  \centering
  % \vspace{-20pt}  % adjust vertical position
  \includegraphics[width=0.30\textwidth]{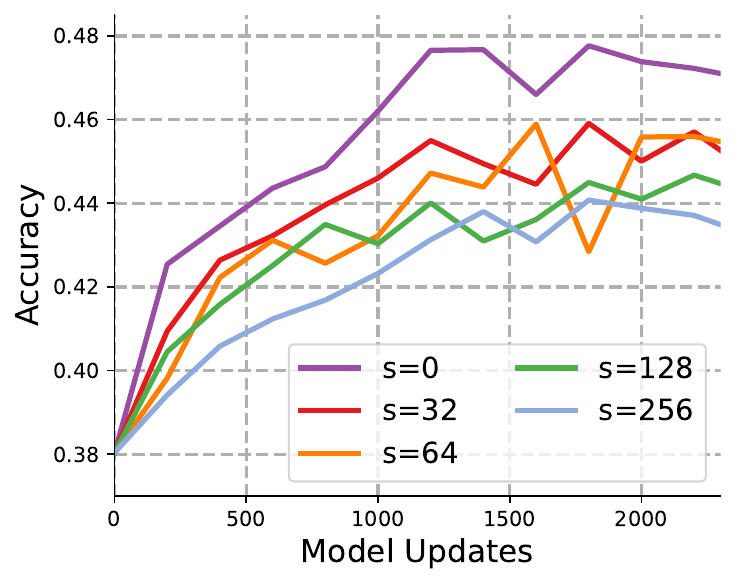}
  \vspace{-8pt}
  \caption{Average accuracy for RL with stale data on Qwen2.5-Math-7B.
  }
  \label{fig:staleness} 
\end{wrapfigure}
\textbf{Stale-$k$ RL training.} 
To investigate the impact of stale data on reinforcement learning for large language models, we introduce Stale-$k$ RL training, where the model is trained using data generated $k$ model updates earlier in each training iteration. 
More specifically, in our training setup, each training step consists of four model updates, a configuration commonly used in recent work~\citep{zheng2025act, wang2025reinforcement, yu2025dapo, chen2025minimax, zheng2025group}.
Thus, even stale-0~($s$=0) training has a staleness between 0 and 3. stale-256~($s$=256) training has a staleness between 256 and 259).
During the first $k$ model updates, since no stale model is yet available, the model is trained on data generated by the original base model, with different training data used in each iteration.
In this setup, all training data after the initial phase comes from stale models, allowing us to study how stale data affects the dynamics and effectiveness of RL training.
More training details can be found at Appendix~\ref{app:setting}.

As shown in Figure~\ref{fig:staleness}, we train Qwen2.5-Math-7B~\citep{yang2024qwen2} with GRPO under varying staleness levels and report test accuracy. The results reveal a clear trend: as staleness increases, model performance degrades and convergence slows. In particular, low-staleness training achieves higher accuracy, whereas high-staleness training converges more slowly to lower performance.
% \vspace{-0.4cm}
\section{Prosperity before Collapse: Stale Data Contain Enough Training Information in RL on LLMs}
\label{sec:pbc}
% \vspace{-0.3cm}

In this section, we investigate why RL on LLM deteriorates when trained on stale data generated by earlier policies.
First, we reveal an intriguing \emph{prosperity-before-collapse} phenomenon: although off-policy RL training without a trust region eventually collapses on stale data, it achieves substantially higher performance than GRPO with $\epsilon$-clipping before collapse, even matching on-policy results.
Next, we study the causes of GRPO’s inferior performance when trained with stale data.

% Interestingly, the results show the opposite of what one might expect: even with highly stale trajectories, training without clipping can initially achieve performance comparable to on-policy baselines, before ultimately collapsing due to uncontrolled variance.

% \subsection{Prosperity before Collapse: Stale data contain enough training information}
\label{ssec:prosperity}
% \label{ssec:pivotal}

\begin{wrapfigure}{r}{0.53\textwidth}
  \centering
  % \vspace{-10pt}  % adjust vertical position
  \includegraphics[width=0.53\textwidth]{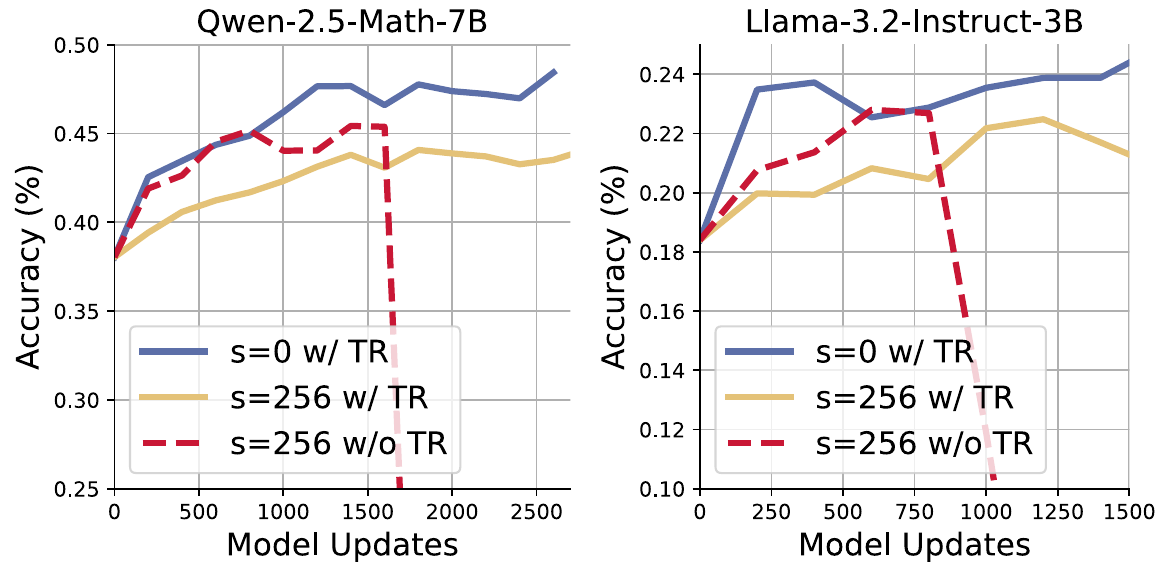}
  % \vspace{-10pt}
\caption{Prosperity before Collapse. Training without a trust region~(TR) ($\epsilon=\infty$) under stale data ($s=256$) initially achieves higher performance than clipped training, sometimes even matching the on-policy baseline ($s=0$). However, it eventually collapses due to uncontrolled variance.}
  \label{fig:pbc} 
\end{wrapfigure}
\textbf{Prosperity before collapse: training without a trust region.}
To disentangle whether the performance drop stems from stale data generated by highly shifted old policies or from biases introduced by the training algorithm, we remove the trust region entirely to remove bias from the training algorithm. 
Surprisingly, we observe a distinct \emph{prosperity-before-collapse} phenomenon. As shown in Figure~\ref{fig:fig1} and Figure~\ref{fig:pbc}, although training without a trust region eventually collapses, it achieves substantially better performance prior to collapse. 
In fact, under stale data ($s$=256), the no-clipping setting outperforms clipped training, sometimes even matching on-policy baselines.

\begin{wrapfigure}{r}{0.55\textwidth}
    \centering
    % \vspace{-10pt}
    \subfloat[]{
        \includegraphics[width=0.255\textwidth]{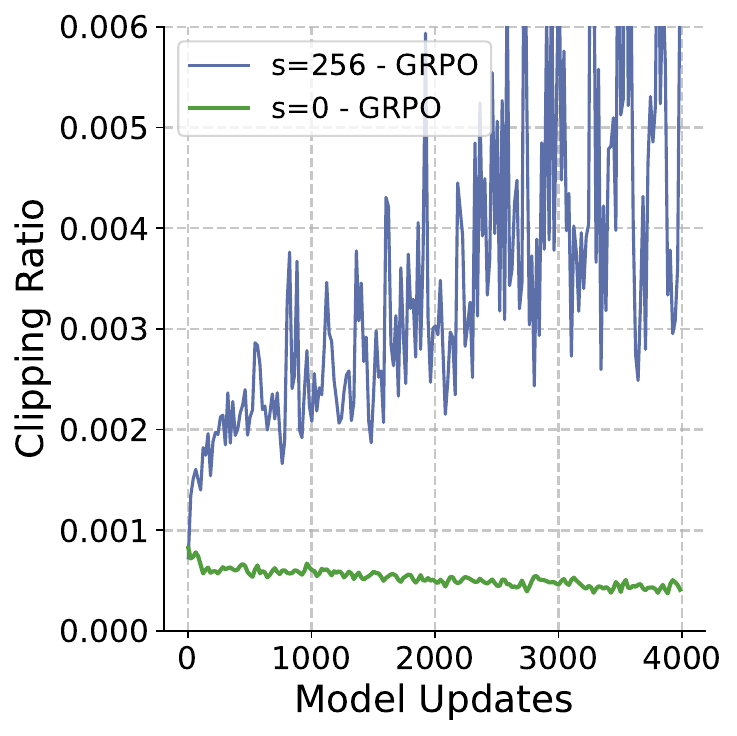}
        \label{fig:clipping-7b}
    }
    \hfill
    \subfloat[]{
        \includegraphics[width=0.255
        \textwidth]{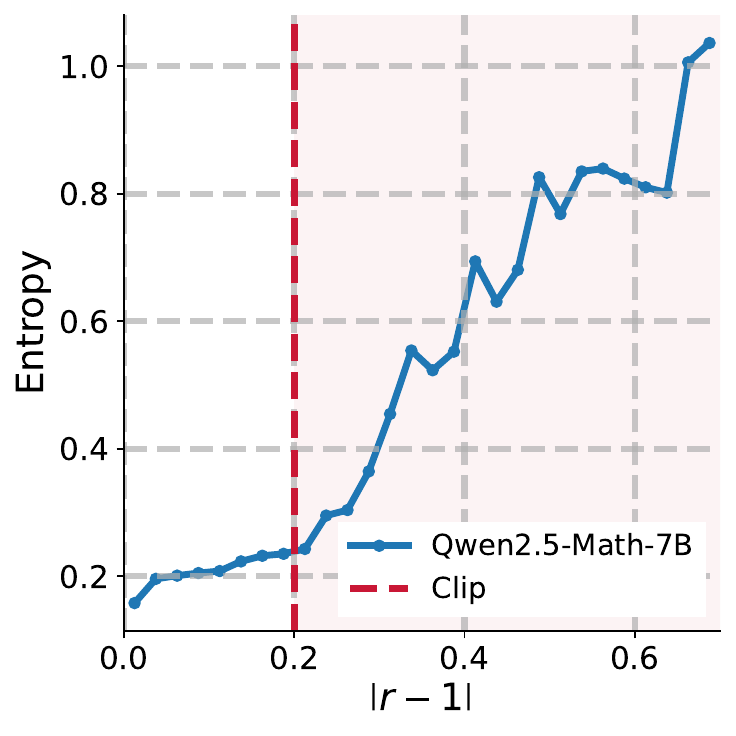}
        \label{fig:entropy-ratio}
    }
    \caption{ 
    \textbf{(a)} Clipping ratio during training on the Qwen-2.5-Math-7B model.
    \textbf{(b)} Relationship between average token entropy and the distance between the importance sampling ratio and 1.}
    \label{fig:token-masking}
    % \vspace{-5pt} 
\end{wrapfigure}

\textbf{Pivotal token masking by $\epsilon$-clipping when training with stale data.}
As also discussed in recent work~\citep{su2025klear, chen2025minimax}, $\epsilon$-clipping may inadvertently mask important tokens, preventing them from contributing useful training signals. 
We extend this observation to the asynchronous setting and show that the problem becomes substantially more severe when training with stale data, since larger staleness induces a greater mismatch between the behavior and target policies. As illustrated in Figure~\ref{fig:clipping-7b}, the clipping ratio increases sharply under large staleness ($s=256$), while remaining much lower in the on-policy baseline.

To better understand this phenomenon, we conduct a quantitative analysis on 90 million training tokens collected during Qwen2.5-Math-7B training with staleness $256$. Specifically, we gather all training tokens generated between 800 and 1200 model updates, ensuring the model is already in a stable training phase but before convergence. Figure~\ref{fig:entropy-ratio} shows a clear trend: as $|r-1|$ increases, the average token entropy also rises. This indicates that $\epsilon$-clipping disproportionately prunes high-entropy tokens, which are typically the most informative for model improvement. Consequently, clipping under stale data leads to degraded performance. 

This observation reveals a dilemma: while high-entropy tokens are crucial for learning progress, they also introduce instability in the off-policy setting, which motivates our key research question:

\begin{center}
\begin{tcolorbox}[width=0.85\linewidth]
\textit{Can a more accurate and adaptive trust region strategy preserve the benefits of stale data while ensuring stable training?}
\end{tcolorbox}
\end{center}

\section{Second-Moment Trust Policy Optimization}
\setlength{\textfloatsep}{8pt}    % 算法与正文之间的间距
\setlength{\floatsep}{6pt}        % 两个 float 之间的间距
\setlength{\intextsep}{6pt}       % 插入在正文中的 float 的上下间距

In this section, we propose Second-Moment Trust Policy Optimization~(M2PO), a novel policy optimization algorithm providing a more effective trust region for off-policy training with stale data.
As discussed in Section~\ref{sec:pbc}, as high-entropy tokens are a double-edged sword, the key challenge in designing an effective trust region algorithm is how to best harness the rich information in high-entropy tokens without letting them destabilize training.

\subsection{Measuring Distribution Gap with the Second Moment}

The main source of instability in off-policy RL lies in the distributional mismatch between the behavior policy that generates training data and the current policy being optimized~\citep{schulman2015trust, schulman2017proximal}. As the divergence between these two distributions grows, importance sampling corrections produce high-variance gradient estimates, leading to noisy and unreliable updates.
Our motivation is therefore to constrain the distributional gap between $\pi_{\text{behav}}$ and $\pi_\theta$ at the batch level, directly coupling the constraint with model updates while preventing over-constraining of token-level variations.

A natural choice to measure distribution is the batch-level KL divergence, a metric widely adopted to monitor stability in RL:

\begin{equation}
\hat{KL} = \frac{1}{N} \sum_{i=1}^{N}\hat{KL}_i 
= -\frac{1}{N} \sum_{i=1}^{N} \log r_i
= -\frac{1}{N} \sum_{i=1}^{N} \log \frac{\pi_{\theta}(a_i \mid s_i)}{\pi_{\text{behav}}(a_i \mid s_i)},
\end{equation}

where $N$ is the number of tokens in a batch.

However, batch-level KL suffers from two key limitations. 
First, because it is computed from single-sample estimates, individual $\hat{KL}_i$ can be positive or negative, leading to \textit{cancellation effects} where large deviations offset each other and produce deceptively small KL values. 
Second, tokens with large ratios ($r_i>1$) are not properly constrained, as their negative $\hat{KL}_i$ actually decreases the estimated KL, even though such tokens can contribute to training instability~\citep{schulman2017proximal}.

To overcome these limitations, we propose to use the second moment of the log-ratio to measure the distribution gap between behavior and current policy. 
Formally, we define

\begin{equation}
\hat{M}_2 = \frac{1}{N} \sum_{i=1}^{N}\hat{M}_{2,i} 
= \frac{1}{N} \sum_{i=1}^{N} (\log r_i)^2
=\frac{1}{N} \sum_{i=1}^{N} [\log \frac{\pi_{\theta}(a_i \mid s_i)}{\pi_{\text{behav}}(a_i \mid s_i)}]^2,
\end{equation}

This choice is motivated by two key advantages of $\hat{M}_2$ over the batch $\hat{KL}$. 
First, each per-token estimate $\hat M_{2,i} = (\log r_i)^2$ is always non-negative, 
so the constraint can be reliably applied even when $r>1$. 
Second, while the batch KL only measures the mean shift between policies, 
$M_2$ also reflects the variance of importance weights. 
This makes $\hat{M}_2$ more sensitive to outliers and noisy tokens with extreme ratios $r_i$\footnote{
A potential alternative is to use $\sum_{i=1}^{N}|\hat{KL}_i| / N$. While this absolute KL estimate can also work empirically, it is less sensitive to variance compared to $M_2$
Moreover, M2 provides an upper bound for this absolute KL estimate, as $E[|r|] \;\leq\; \sqrt{E[r^2]}$. Therefore, we adopt $M_2$ in our method.
}.

Furthermore, Theorem \ref{thm:bound} shows that although $M_2$ does not directly constrain $r-1$ like $\epsilon$ clipping, it nevertheless provides an upper bound on the Pearson chi-square divergence $\mathbb{E}[(r-1)^2]$ between the new and behavior policies. The proof is provided in Appendix~\ref{app:proof}.

% \clearpage

\begin{theorem}[Bounding $\chi^2$ by $M_2$]
\label{thm:bound}
Let $r=\tfrac{\pi_{\mathrm{new}}}{\pi_{\mathrm{behav}}}$ be the importance ratio and assume $1/R \le r\le R$. 
Define the log-ratio second moment
\[
M_2 = \mathbb{E}_{a\sim \pi_{\mathrm{behav}}}\!\big[(\log r(a))^2\big].
\]
Let the Pearson chi-square divergence between $\pi_{\mathrm{new}}$ and $\pi_{\mathrm{behav}}$ be
\[
\chi^2(\pi_{\mathrm{new}}\!\parallel\!\pi_{\mathrm{behav}}) 
= \mathbb{E}_{a\sim \pi_{\mathrm{behav}}}\!\left[\left(\frac{\pi_{\mathrm{new}}(a)}{\pi_{\mathrm{behav}}(a)}-1\right)^2\right]
= \mathbb{E}_{\pi_{\mathrm{behav}}}\!\big[(r-1)^2\big].
\]
Then
\[
\chi^2(\pi_{\mathrm{new}}\!\parallel\!\pi_{\mathrm{behav}})\;\le\; R^2\, M_2.
\]
\end{theorem}

\subsection{Second-Moment Trust Policy Optimization}

\begin{wrapfigure}{r}{0.45\textwidth} % r=右侧，l=左侧
% \vspace{-1em}
  \newlength{\wrappad}\setlength{\wrappad}{0.5em} % desired extra left space
  \hspace{\wrappad}%
\begin{minipage}{0.4\textwidth}
\begin{algorithm}[H]
\caption{M2PO Masking}
\label{alg:m2po}
\KwIn{$\{\hat M_{2,i}\}_{i=1}^N$ for all training tokens; threshold $\tau_{M_2}$}
\KwOut{mask $\boldsymbol M$}
$\boldsymbol M \gets \texttt{True}$ for all tokens; \\
$\mathcal T \gets$ all trust-region tokens; \\
\While{$\operatorname{mean}_{i\in\mathcal T}\hat M_{2,i} \;>\; \tau_{M_2}$}{
  $j \gets \arg\max_{i\in\mathcal T}\hat M_{2,i}$; \\
  $\boldsymbol M_j \gets \texttt{False}$; \quad $\mathcal T \gets \mathcal T \setminus \{j\}$;
}
\Return $\boldsymbol M$
\end{algorithm}
\vspace{-1em}
\end{minipage}
\end{wrapfigure}
As illustrated in Algorithm~\ref{alg:m2po}, to maintain training stability, M2PO applies a masking strategy that selectively excludes tokens until the batch-level $\hat{M}_2$ of the remaining tokens falls below a predefined threshold $\tau_{M_2}$. 
Importantly, we observe that $\tau_{M_2}$ is not a sensitive hyperparameter (see Figure~\ref{fig:ablation-m2}). 
Across all our experiments, we consistently set $\tau_{M_2} = 0.04$, and this single setting proved effective for stabilizing training in all training scenarios.

\textbf{Only constrain trust-region tokens.} 
Although the PPO loss clips the ratio on both the upper and lower sides, 
due to the use of the \texttt{min} operator, not all tokens are actually clipped. 
In practice, clipping only occurs for tokens where 
$A > 0$ and $r > 1$, or $A < 0$ and $r < 1$. 
Following the PPO setting, we therefore apply the $M_2$ constraint exclusively 
to tokens that satisfy these conditions.
Finally, with the result mask $\boldsymbol{M}$, we update the policy by maximizing the following objective\footnote{In our loss, we average over all tokens rather than only the unmasked ones. 
This choice is intended to better mimic the behavior of PPO-style clipping. However, since the masking ratio is typically very small (see Section~\ref{ssec:analysis}), the difference between the two averaging strategies is negligible in practice.}:

\begin{equation}
\mathcal{J}_{\text{M2PO}}(\theta)
=
\frac{1}{\sum_{i=1}^G |o_i|}
\sum_{i=1}^{G}\sum_{t=1}^{|o_i|}
\boldsymbol{M}_{i,t}\, \frac{\pi_\theta(o_i |q)}{\pi_{\theta_{behav}}(o_i |q)}\,A_{i,t},
\qquad
\boldsymbol{M}_{i,t}\in\{0,1\},
\end{equation}

where $A_i$ denotes the advantage, computed using the grouped advantage in Equation~\ref{eq:group-adv}.

% \vspace{-0.2cm}
\section{Experiments}
\label{sec:exp}

In this section, we present an extensive evaluation across six models~(from 1.7B to 32B) on eight benchmarks. The results demonstrate that, even when trained with extremely stale data, M2PO achieves performance comparable to on-policy GRPO and significantly outperforms other baselines:

\begin{denseitemize}

 \item In Section~\ref{ssec:performance}, we show that M2PO achieves \textbf{accuracy on par with on-policy baselines} under large staleness ($s=256$), and outperforms baselines by up to 11.2\% in average accuracy.

 \item In Section~\ref{ssec:analysis}, we provide a detailed analysis of how M2PO boosts off-policy RL performance while preserving training stability. We also show that its sole threshold hyperparameter $\tau_{M_2}$ is insensitive to variation, ensuring ease of use in practice.

\end{denseitemize}

% \vspace{-0.2cm}
\subsection{Experimental Settings}
% \vspace{-0.2cm}

\textbf{Models \& Datasets.} 
To verify the effectiveness of our method, we extensively evaluate M2PO on six models: 
Qwen2.5-Math-7B~\citep{yang2024qwen2}, 
Llama-3.2-3B-Instruct~\citep{dubey2024llama}, 
Qwen3-Base-1.7B/4B/8B~\citep{yang2025qwen3}, 
and Qwen2.5-32B~\citep{yang2024qwen2}. 
For Qwen2.5-Math-7B, we use a context length of 4k, which is the maximum for this series, 
while for all other models the context length is set to 16k. 
For training, we adopt the DeepScaleR~\citep{deepscaler2025} math dataset.

\textbf{Training \& Evaluation.} Our method is implemented based on verl~\citep{sheng2024hybridflow} pipeline and uses vLLM~\citep{kwon2023efficient} for rollout. We use a mix of H100 and H200 servers for training, depending on resource availability.
For benchmark datasets, we use eight widely used complex mathematical reasoning benchmarks to evaluate the performance of trained models:
Math500~\citep{hendrycksmath2021, lightman2023lets}, 
AIME24/25~\citep{AoPS:AIMEProblemsSolutions},
AMC23/24~\citep{AoPS:AMCProblemsSolutions},
Minerva Math~\citep{lewkowycz2022solving},
Gaokao~\citep{zhang2023evaluating},
Olympiad Bench~\citep{he2024olympiadbench}.
Similar to ~\citep{wang2025reinforcement, zheng2025act}, we evaluate models on those benchmarks every $50$ steps and report the performance of the checkpoint that obtains the best average performance on eight benchmarks. 
For GRPO, we adopt the commonly used clipping parameter $\epsilon = 0.2$, while for the other baselines, we follow the recommended values reported in their respective papers.
We include more detailed experimental settings in Appendix~\ref{app:setting}.

% \textbf{Baselines.}
% \vspace{-0.2cm}
\subsection{Performance Comparison on Training with Staleness}
\label{ssec:performance}
% \vspace{-0.2cm}

\begin{table}[!t]
\centering
\setlength{\tabcolsep}{4pt}
\caption{Performance (\%) comparison across eight math reasoning benchmarks using models from 1.7B to 32B parameters. We report results for GRPO, GSPO, and M2PO under both on-policy ($s=0$) and off-policy ($s=256$) settings. 
\underline{Underlined} numbers denote the best average accuracy, while \textbf{bold} numbers highlight the best average accuracy under stale rollouts ($s=256$). 
M2PO consistently improves stability under staleness and achieves higher average accuracy than GRPO.}

\label{tab:performance-compare}
\small
\begin{tabular}{c|c|c c c c c c|c}
\toprule
\textbf{Method} & \textbf{S} & \textbf{AIME24/25} & \textbf{AMC23/24} & \textbf{Math500} & \textbf{Gaokao} & \makecell{\textbf{Miner.}} & \makecell{\textbf{Olymp.}} & \textbf{Avg.} \\
\bottomrule
\toprule
\multicolumn{9}{c}{\textbf{\textit{Llama-3.2-3B-Instruct}}} \\
\midrule
GRPO & 0   & 11.0 / 2.3 & 31.3 / 17.2 & 53.6 & 42.99 & 23.1 & 20.3 & 25.2 \\
GRPO & 256 & 9.6 / 0.4  & 25.0 / 13.9 & 52.4 & 42.08 & 17.6 & 18.8 & 22.5 \\
GSPO & 256 & 9.0 / 0.2  & 30.0 / 14.4 & 50.6 & 40.65 & 18.8 & 17.3 & 22.6 \\
M2PO~(Ours) & 256 & 10.4 / 4.4 & 33.8 / 17.8 & 52.0 & 44.48 & 21.2 & 18.1 & \textbf{\underline{25.3}} \\
\midrule
\multicolumn{9}{c}{\textbf{\textit{Qwen2.5-Math-7B}}} \\
\midrule
GRPO & 0   & 39.6 / 17.5 & 63.8 / 46.7 & 82.3 & 64.1 & 36.7 & 43.6 & \underline{49.3} \\
GRPO & 256 & 29.4 / 12.9 & 64.4 / 39.4 & 80.5 & 63.2 & 33.1 & 43.1 & 45.7 \\
GSPO & 256 & 27.3 / 13.1 & 63.8 / 36.7 & 79.0 & 62.2 & 33.5 & 41.9 & 44.7 \\
M2PO~(Ours) & 256 & 33.3 / 17.5 & 63.8 / 40.6 & 84.0 & 66.4 & 38.1 & 47.1 & \textbf{48.8} \\
\midrule
\multicolumn{9}{c}{\textbf{\textit{Qwen3-Base-1.7B}}} \\
\midrule
GRPO & 0   & 7.5 / 7.5   & 40.6 / 26.1 & 67.2 & 55.9 & 28.9 & 30.5 & 33.0 \\
GRPO & 256 & 8.5 / 4.8 & 34.4 / 25.0 & 64.3 & 52.7 & 26.0 & 27.6 & 30.4 \\
GSPO & 256 & 6.9 / 4.0   & 39.4 / 18.9 & 65.0 & 53.1 & 26.5 & 27.5 & 30.1 \\
M2PO~(Ours) & 256 & 14.0 / 6.5  & 48.1 / 27.8 & 71.8 & 59.5 & 29.4 & 35.6 & \textbf{\underline{36.6}} \\
\midrule
\multicolumn{9}{c}{\textbf{\textit{Qwen3-Base-4B}}} \\
\midrule
GRPO & 0   & 22.9 / 20.2 & 63.8 / 53.9 & 84.6 & 69.8 & 40.2 & 50.5 & 50.7 \\
GRPO & 256 & 14.0 / 9.6  & 51.9 / 32.8 & 76.8 & 61.7 & 34.4 & 39.8 & 40.1 \\
GSPO & 256 & 17.9 / 15.4 & 55.6 / 38.3 & 76.8 & 62.3 & 35.1 & 44.3 & 43.2 \\
M2PO~(Ours) & 256 & 26.7 / 21.0 & 64.4 / 49.4 & 85.8 & 70.3 & 40.5 & 52.3 & \textbf{\underline{51.3}} \\
\midrule
\multicolumn{9}{c}{\textbf{\textit{Qwen3-Base-8B}}} \\
\midrule
GRPO & 0   & 26.7 / 19.4 & 76.9 / 52.8 & 87.7 & 71.6 & 41.2 & 52.8 & 53.6 \\
GRPO & 256 & 21.0 / 13.1 & 63.8 / 40.0 & 81.8 & 67.8 & 38.5 & 47.4 & 46.7 \\
M2PO~(Ours) & 256 & 30.2 / 23.1 & 71.3 / 56.7 & 87.2 & 75.1 & 42.6 & 54.8 & \textbf{\underline{55.1}} \\
\midrule
\multicolumn{9}{c}{\textbf{\textit{Qwen2.5-32B}}} \\
\midrule
GRPO & 0   & 24.4 / 18.3 & 71.9 / 46.7 & 85.4 & 71.9 & 41.4 & 52.9 & 51.6 \\
GRPO   & 256 & 20.4 / 9.6  & 68.1 / 41.1 & 83.0 & 67.3 & 40.9 & 45.9 & 47.0 \\
M2PO~(Ours)   & 256 & 24.8 / 19.4 & 76.3 / 50.0 & 85.7 & 71.7 & 41.5 & 51.7 & \textbf{\underline{52.6}} \\
\bottomrule
\end{tabular}
\end{table}

\textbf{Prosperity without collapse: Stable off-policy training without performance degradation using M2PO.}
To verify the effectiveness of M2PO, Table~\ref{tab:performance-compare} presents a comprehensive comparison of math reasoning performance across eight benchmarks using models from four different families and scales, ranging from 1.7B to 32B parameters. We evaluate multiple reinforcement learning methods under both on-policy and off-policy settings, including GRPO, GSPO, and our proposed M2PO.
The results show that while 
\begin{wrapfigure}{r}{0.36\textwidth}
  \centering
  % \vspace{-20pt}  % adjust vertical position
  \includegraphics[width=0.34\textwidth]{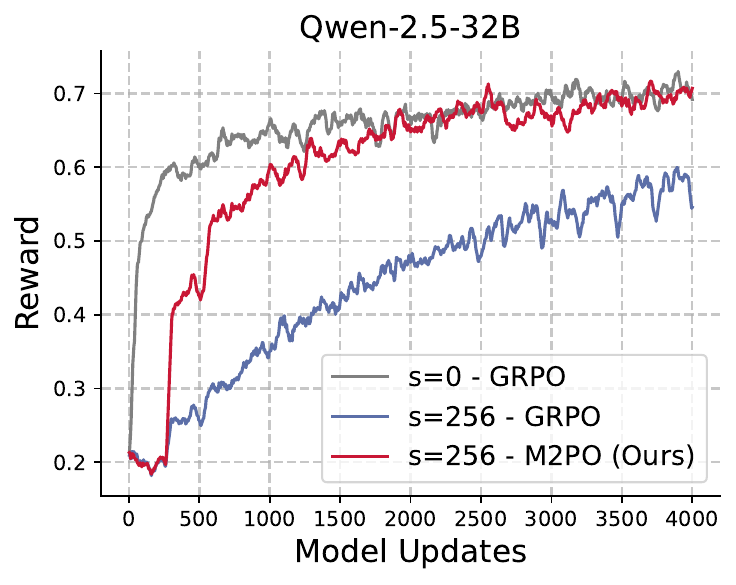}
  % \vspace{-10pt}
  \caption{Training reward on Qwen-2.5-32B.
  }
  % \vspace{-0.1cm}
  \label{fig:reward-32b} 
\end{wrapfigure}
both GRPO and GSPO often suffer significant performance drops under large staleness, M2PO consistently achieves comparable accuracy to the on-policy baseline in all training settings.
Surprisingly, we notice that, in some model settings, M2PO with $s=256$ even achieves a better performance than M2PO with $s=0$.
For instance, on the Qwen3-Base-1.7B model, we observe that M2PO with $s=256$ (36.6\%) outperforms GRPO with $s=0$ (33.0\%).
\begin{wrapfigure}{r}{0.36\textwidth}
  \centering
  % \vspace{-20pt}  % adjust vertical position
  \includegraphics[width=0.34\textwidth]{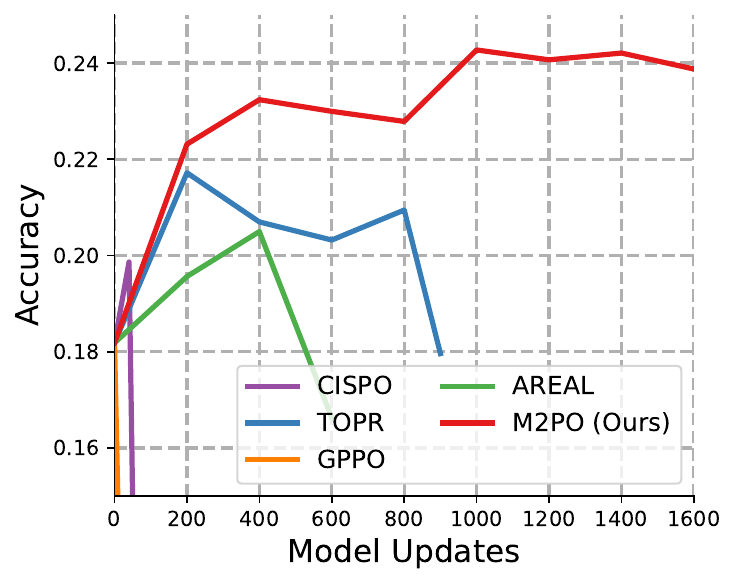}
  % \vspace{-10pt}
  \caption{Methods comparison under staleness ($s=256$) on Llama-Instruct-3B.
  }
  \label{fig:other-baselines} 
\end{wrapfigure}
A potential explanation is that small effective staleness (e.g., $s=0$ corresponding to delays between 0 and 3) can still adversely affect training stability.
Our further analysis in Figure~\ref{fig:clipping} supports this view, showing that M2PO with $s=256$ exhibits an even lower clipping ratio than GRPO with $s=0$.
Overall, these results show that M2PO remains robust and effective, sustaining stable, high performance even under extreme off-policy conditions.

In addition to the final accuracy comparison in Table~\ref{tab:performance-compare}, we also analyze the training dynamics of accuracy and reward of Qwen-2.5-32B models.
As shown in Figure~\ref{fig:fig1}, 
M2PO with $s=256$ initially falls behind the on-policy baseline but quickly catches up, eventually matching its performance, while converging much faster and achieving higher accuracy than GRPO under the same staleness.
This highlights that M2PO not only maintains comparable final accuracy but also accelerates convergence when training with stale data.
Figure~\ref{fig:reward-32b} shows a similar trend in the reward curves. 
M2PO with $s=256$ also starts off behind the on-policy baseline due to the initial plateau caused by using data generated from the base model, but it quickly catches up and aligns closely with the $s=0$ trajectory. 
In contrast, GRPO with $s=256$ consistently underperforms across the entire training trajectory.

% AREAL~\citep{fu2025areal}, for instance, relies on an approximate policy to define the trust region, but this approximation can become inaccurate when the behavior policy drifts too far under large staleness.
\textbf{Performance of other baselines under staleness.}
A number of prior works have proposed alternative trust region strategies beyond $\epsilon$-clipping, including GSPO~\citep{zheng2025group}, AREAL~\citep{fu2025areal}, TOPR~\citep{roux2025tapered}, GPPO~\citep{su2025klear}, and CISPO~\citep{chen2025minimax}. 
\emph{Despite not being designed to handle the extreme staleness studied in this paper}, we also evaluate these methods under our setting with $s=256$ for completeness and comparison. 
Among these methods, GSPO is the only one that preserves training stability, though it still exhibits a noticeable performance drop under high staleness~(see Table~\ref{tab:performance-compare}).
For the other methods, as shown in Figure~\ref{fig:other-baselines}, most encounter substantial difficulties in maintaining training stability and tend to break down early in training.
These observations suggest that while existing approaches can be effective under moderately stale settings, they face significant challenges when extended to larger staleness, highlighting the need for a more robust and effective solution.

\subsection{Analysis and Ablation Study}
\label{ssec:analysis}

\begin{wrapfigure}{r}{0.55\textwidth}
    \centering
    % \vspace{-10pt}
    \vspace{-0.9cm}
    \subfloat[]{
        \includegraphics[width=0.255\textwidth]{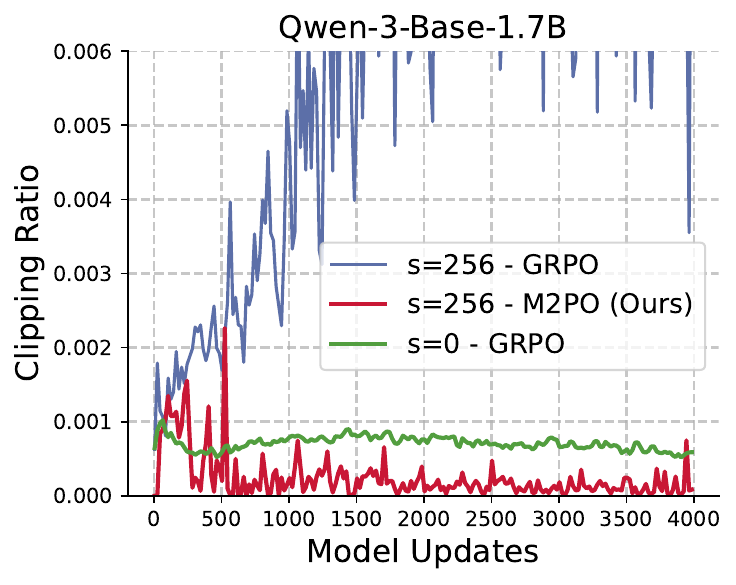}
        \label{fig:clipping-17b}
    }
    \hfill
    \subfloat[]{
        \includegraphics[width=0.255\textwidth]{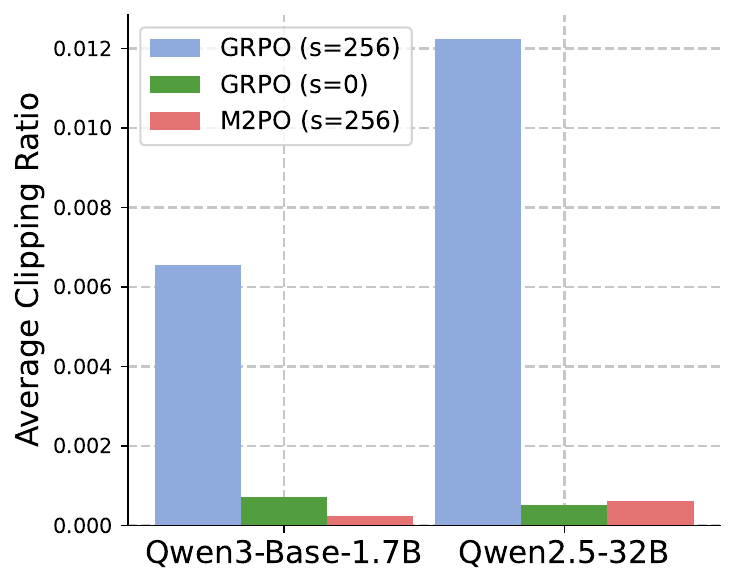}
        \label{fig:clipping-bar}
    }
    \caption{%
    \textbf{(a)} Clipping ratio dynamics during RL on the Qwen-3-Base-1.7B model. 
    \textbf{(b)} Comparison of the average clipping ratio across models and methods.}
    \vspace{-0.2cm}
    \label{fig:clipping}
\end{wrapfigure}
\textbf{Stable training with reduced clipping.}
Figure~\ref{fig:clipping}(a)(b) illustrates the clipping dynamics of GRPO and our proposed M2PO under different staleness settings.
In Figure~\ref{fig:clipping-17b}, we report results on Qwen-3-Base-1.7B. Under large staleness ($s=256$), GRPO exhibits frequent clipping events, with the ratio increasing sharply and remaining high during most of the training. 
In contrast, M2PO under the same staleness maintains an exceptionally low clipping ratio, comparable to or even lower than the on-policy GRPO baseline ($s=0$). 
Notably, M2PO with $s=256$ exhibits less clipping than GRPO with $s=0$, which explains why M2PO with $s=256$ achieves higher accuracy than GRPO with $s=0$ in Table~\ref{tab:performance-compare}.
Figure~\ref{fig:fig1} shows the same comparison on Qwen-2.5-32B. A similar trend holds: GRPO with $s=256$ suffers from substantial clipping, whereas M2PO effectively suppresses unnecessary clipping, remaining close to the on-policy baseline.

\begin{wrapfigure}{r}{0.32\textwidth}
  \centering
  % \vspace{-20pt}  % adjust vertical position
  \includegraphics[width=0.315\textwidth]{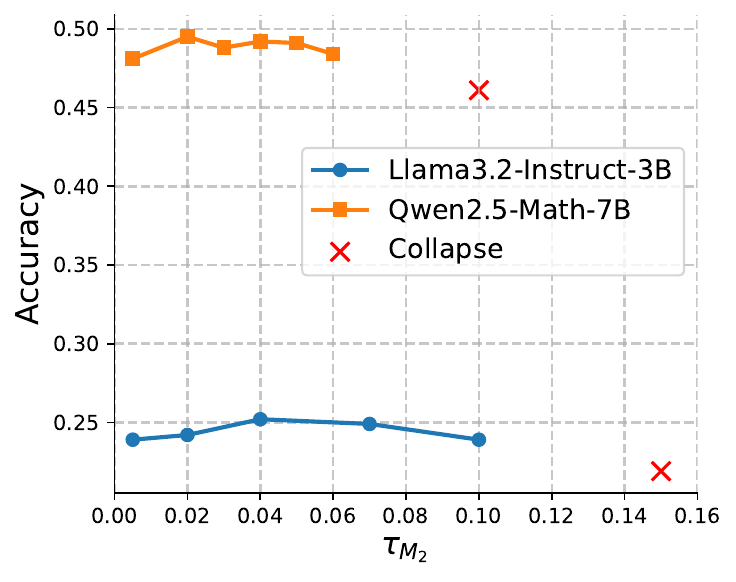}
  % \vspace{-10pt}
  \caption{Ablation study of the $\tau_{M_2}$ threshold on Llama-3.2-3B-Instruct and Qwen2.5-Math-7B.}
  \label{fig:ablation-m2} 
\end{wrapfigure}
Figure~\ref{fig:clipping-bar} summarizes the average clipping ratio across the entire training process. On Qwen-3-Base-1.7B, GRPO with $s=256$ reaches an average clipping ratio of 0.66\%, compared to 0.07\% for GRPO with $s=0$ and only 0.02\% for M2PO with $s=256$. On Qwen-2.5-32B, GRPO with $s=256$ averages 1.22\%, while GRPO with $s=0$ records 0.05\% and M2PO with $s=256$ maintains a similarly low ratio of 0.06\%. 
These results show that M2PO reduces clipping by over an order of magnitude compared to GRPO, thereby enabling stable and efficient training by clipping only when necessary.

\textbf{Robustness to the choice of $\tau_{M_2}$.}
Figure~\ref{fig:ablation-m2} shows that performance is not highly sensitive to the choice of $\tau_{M_2}$. Accuracy remains stable across a broad range, only dropping when $\tau_{M_2}$ is set extremely small (overly restrictive constraint) or very large (training collapse). This explains why a single setting of $\tau_{M_2}=0.04$ works robustly across all training settings in our paper.

\begin{wrapfigure}{r}{0.55\textwidth}
    \centering
    % \vspace{-10pt}
    \subfloat[]{
        \includegraphics[width=0.255\textwidth]{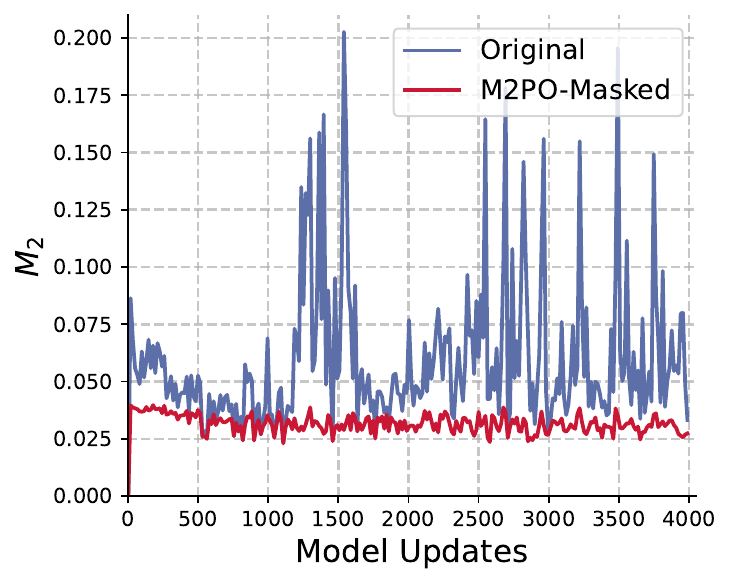}
        \label{fig:m2-dynamics}
    }
    \hfill
    \subfloat[]{
        \includegraphics[width=0.255\textwidth]{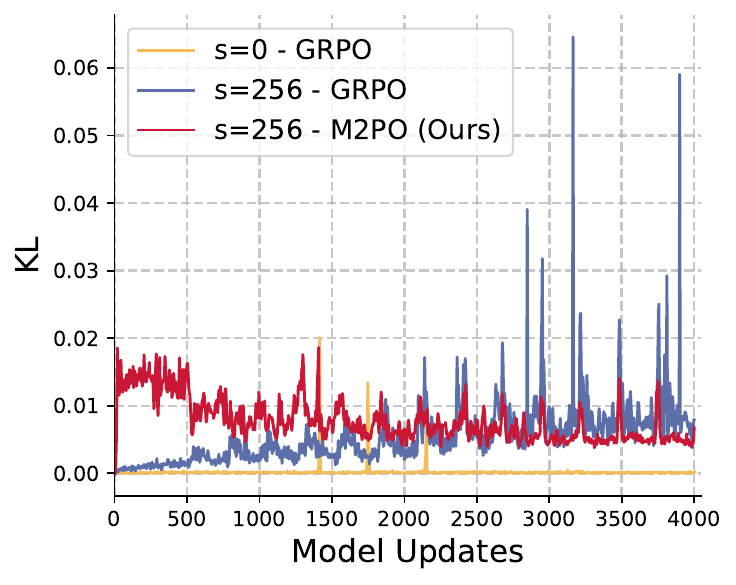}
        \label{fig:kl-dynamics}
    }
    \caption{ 
    (a) Average $M_{2}$ in each model updates with and without M2PO masking on Qwen-2.5-32B, showing that masking effectively suppresses spikes and stabilizes the $M_{2}$ throughout training. 
    (b) Average KL divergence in each model updates on Qwen-2.5-32B under different methods. 
    }
    \label{fig:m2-kl-dynamics}
    % \vspace{-5pt} 
\end{wrapfigure}
\textbf{Training dynamics on KL and $\mathbf{M_{2}}$.} 
Figure~\ref{fig:m2-kl-dynamics} shows the impact of M2PO masking on training stability. 
Figure~\ref{fig:m2-dynamics} shows that the average $M_{2}$ without masking exhibits frequent spikes throughout training, indicating instability in the second-moment estimates~(blue curve). 
Applying M2PO masking effectively suppresses these fluctuations and maintains consistently low $M_{2}$ values, leading to more stable updates~(red curve). 
Figure~\ref{fig:kl-dynamics} compares the KL divergence across different methods. 
Although M2PO with $s=256$ involves substantially less clipping than GRPO~(shown in Figure~\ref{fig:clipping}), it maintains a more stable divergence than GRPO with $s=256$.
These results indicate that M2PO performs clipping in a more precise and adaptive manner, ensuring training stability with substantially less reliance on clipping.
These results demonstrate that M2PO enables more precise and adaptive clipping, achieving training stability while relying on significantly fewer clipping operations, and thereby attaining better performance without risking training collapse.

% \begin{wrapfigure}{r}{0.4\textwidth}
%   \centering
%   \includegraphics[width=0.39\textwidth]{figs/clipping-token-examples.png}
%   \caption{Word clouds of frequently clipped tokens with $\epsilon$ clipping.}
%   \label{fig:clipped-tokens} 
% \end{wrapfigure}
% \textbf{Commonly Clipped Tokens in GRPO.}
% Figure~\ref{fig:clipped-tokens} shows the specific tokens that are most frequently clipped by $\epsilon$-clipping. 
% The word cloud of commonly clipped token is highly aligned with high-entropy tokens shown in \citep{wang2025beyond}: these tokens are not random or unimportant, but rather belong to the most semantically and structurally critical elements in reasoning traces. Many of them (e.g., First, simplify, determine, To def, Thus, verify, break) are precisely the high-entropy “pivotal tokens” that initiate, connect, or conclude key reasoning steps. Others (e.g., assistant, user, code markers like \#\#\# or \$\$) serve as structural anchors in the dialogue or mathematical formatting.
% This observation aligns with Figure~\ref{fig:entropy-ratio}: as the importance weight ratio $|r-1|$ grows, clipped tokens tend to exhibit higher entropy.

\section{Conclusion}

In this work, we investigated why off-policy RL for LLMs often fails under stale data and uncovered the \emph{prosperity-before-collapse} phenomenon: training without a trust region initially outperforms standard methods, showing that stale data can be as informative as on-policy trajectories, but eventually collapses due to instability. Motivated by this observation, we proposed \textbf{M2PO}, which constrains the second moment of importance weights to provide a variance-sensitive and stable trust region. This design suppresses extreme outliers while preserving informative high-entropy tokens, enabling stable training that matches on-policy performance even under extreme staleness. Extensive experiments further demonstrate that M2PO significantly reduces clipping and is highly insensitive to its threshold, highlighting its practicality and scalability for efficient RL with LLMs.

% \input{arXiv-2410.16179v4/ICLR_2025_Template/introduction}
% \input{arXiv-2410.16179v4/ICLR_2025_Template/problemstatement}
% \input{arXiv-2410.16179v4/ICLR_2025_Template/rewriteobservation}
% \input{arXiv-2410.16179v4/ICLR_2025_Template/gsminfinite}
% % \input{arXiv-2410.16179v4/ICLR_2025_Template/benchmarksandresults}
% \input{arXiv-2410.16179v4/ICLR_2025_Template/analysis}
% \input{arXiv-2410.16179v4/ICLR_2025_Template/conclusion}

% \clearpage
% \newpage
\bibliographystyle{assets/plainnat}
\bibliography{paper}

\clearpage
\newpage
\beginappendix

\section*{Acknowledgment}
We would like to thank Cheng Luo, Xinyu Yang, Ranajoy Sadhukhan, Xuesheng Liu, Yongji Wu for providing us constructive feedback on our paper and the computing resources of NVIDIA. 
This work is supported in part by the grants NSF CCF-2504353 to B. Chen.
This work is also partially supported by Google Research Award, Amazon Research Award, Intel, Li Auto, Moffett AI, and CMU CyLab Seed funding.
We are also grateful to BitDeer AI Research for providing GPU resources. 
Any opinions, findings, and conclusions or recommendations expressed are those of the authors and do not necessarily reflect the views of the National Science Foundation.

\section*{Ethics Statement}
This work focuses on developing reinforcement learning algorithms for large language models. Our research does not involve human subjects, personally identifiable information, or sensitive data. All datasets used are publicly available and widely adopted in the community. We acknowledge that more capable LLMs may have potential societal impacts, including misuse for generating misleading or harmful content. To mitigate these risks, our study is confined to controlled academic settings, and our primary goal is to improve the stability and efficiency of training methods.

\section*{The Use of Large Language Models (LLMs)}
We used large language models (LLMs) as general-purpose assistants in two limited ways: (1) for writing polish, including improving grammar, readability, and presentation of the manuscript, and (2) as code assistants (e.g., Cursor, GitHub Copilot) to accelerate routine coding tasks such as debugging syntax errors and refactoring simple functions. LLMs were not used for research ideation, algorithm design, experimental analysis, or drawing conclusions. All conceptual and scientific contributions are entirely the work of the authors.

\section{Overview}
\label{app:overview}

In this appendix, we provide additional details to complement the main text. Appendix~\ref{app:setting} describes the experimental setup in full, including datasets, models, and training hyperparameters. 
Appendix~\ref{app:proof} presents theoretical proofs supporting our method and analysis. 
Appendix~\ref{app:exp} includes additional experimental results.

\section{Detailed Experimental Setting}
\label{app:setting}

\textbf{Models \& Datasets.} 
To verify the effectiveness of our method, we extensively evaluate M2PO on six models from four model series: 
Qwen2.5-Math-7B~\citep{yang2024qwen2}, 
Llama-3.2-3B-Instruct~\citep{dubey2024llama}, 
Qwen3-Base-1.7B/4B/8B~\citep{yang2025qwen3}, 
and Qwen2.5-32B~\citep{yang2024qwen2}. 
For Qwen2.5-Math-7B, we use a context length of 4k, which is the maximum for this series. 
while for all other models the context length is set to 16k. 
For training, we adopt the DeepScaleR~\citep{deepscaler2025} math dataset.

\textbf{Training.} Our method is implemented based on verl~\citep{sheng2024hybridflow} pipeline and uses vLLM~\citep{kwon2023efficient} for rollout.
We use a mix of H100 and H200 servers for training, depending on resource availability.
We set the rollout temperature to $1$ for vLLM~\citep{kwon2023efficient}. 
% For Qwen2.5-Math-1.5B/7B models, we use $4096$ as the context length, as it is the maximum context length for those two models.
% For DeepSeek-R1-Distill-Qwen-1.5B, we set the context length to $8196$.
The training batch size is set to $256$, and the mini-batch size to $512$. 
We sample $8$ responses per prompt.
We train all models for $1000$ steps, 
and we optimize the actor model using the AdamW~\citep{loshchilov2019decoupled} optimizer with a constant learning rate of 1e-6. We use $\beta_1=0.9$, $\beta_2=0.999$, and apply a weight decay of $0.01$. 
We use the following question template to prompt the LLM. For reward assignment, we give a score of 0.1 for successfully extracting an answer and a score of 1.0 if the extracted answer is correct.
Similar to \citep{yu2025dapo}, we remove the KL-divergence term.
The optimization is performed on the parameters of the actor module wrapped with Fully Sharded Data Parallel~(FSDP)~\citep{zhao2023pytorch} for efficient distributed training.
We set the M2PO threshold to 0.04 for all training runs.

\textbf{Evaluation.}
For benchmark datasets, we use eight widely used complex mathematical reasoning benchmarks to evaluate the performance of trained models:
Math500~\citep{hendrycksmath2021, lightman2023lets}, 
AIME24/25~\citep{AoPS:AIMEProblemsSolutions},
AMC23/24~\citep{AoPS:AMCProblemsSolutions},
Minerva Math~\citep{lewkowycz2022solving},
Gaokao~\citep{zhang2023evaluating},
Olympiad Bench~\citep{he2024olympiadbench}.
Same as the training setting, For Qwen2.5-Math-7B models, we use 4k as the context length.
For other models, we set the context length to 16k.
Similar to ~\citep{wang2025reinforcement, zheng2025act}, we evaluate models on those benchmarks every $50$ steps and report the performance of the checkpoint that obtains the best average performance on eight benchmarks.
We evaluate all models with temperature $=1$.
For AIME24/25, we report the  $pass@1 (avg@16)$, for other benchmarks, we report the $pass@1 (avg@4)$.

\begin{tcolorbox}[myexample={Question Template}]
Please solve the following math problem: \{\{Question Description\}\}. The assistant first thinks about the reasoning process step by step and then provides the user with the answer. Return the final answer in \textbackslash boxed\{\} tags, for example \textbackslash boxed\{1\}. Let's solve this step by step. 
\end{tcolorbox}

\section{Theoretical Proof}
\label{app:proof}

% \begin{theorem}[Bounding $\chi^2$ by $M_2$]
% Let $r=\tfrac{\pi_{\mathrm{new}}}{\pi_{\mathrm{behav}}}$ be the importance ratio and assume $1/R \le r\le R$. 
% Define the log-ratio second moment
% \[
% M_2 = \mathbb{E}_{a\sim \pi_{\mathrm{behav}}}\!\big[(\log r(a))^2\big].
% \]
% Let the Pearson chi-square divergence between $\pi_{\mathrm{new}}$ and $\pi_{\mathrm{behav}}$ be
% \[
% \chi^2(\pi_{\mathrm{new}}\!\parallel\!\pi_{\mathrm{behav}}) 
% = \mathbb{E}_{a\sim \pi_{\mathrm{behav}}}\!\left[\left(\frac{\pi_{\mathrm{new}}(a)}{\pi_{\mathrm{behav}}(a)}-1\right)^2\right]
% = \mathbb{E}_{\pi_{\mathrm{behav}}}\!\big[(r-1)^2\big].
% \]
% Then
% \[
% \chi^2(\pi_{\mathrm{new}}\!\parallel\!\pi_{\mathrm{behav}})\;\le\; R^2\, M_2.
% \]
% \end{theorem}

\paragraph{Proof of Theorem~\ref{thm:bound}.}

\begin{proof}
Let $z=\log r$, so that $r=e^Z$. Since $1/R<r\le R$, we have $|z|\le \log R$.  

For $z\ge 0$, observe that
\[
\frac{e^z-1}{z} = \int_0^1 e^{tz}\,dt \;\le\; \int_0^1 e^{z}\,dt = e^z,
\]
which implies $(e^z-1)^2 \le (z e^z)^2 = z^2 e^{2z}$.

For $z\le 0$, set $u=-z\ge 0$. Then
\[
(e^z-1)^2 = (1-e^{-u})^2 \;\le\; u^2 = z^2 \;\le\; z^2 e^{2|z|}.
\]

Combining both cases, for all $z\in\mathbb{R}$ we obtain
\[
(e^z-1)^2 \le z^2 e^{2|z|}.
\]

Substituting $Z=\log r$, this yields
\[
(r-1)^2 \le (\log r)^2 e^{2|\log r|} \;\le\; R^2 (\log r)^2.
\]

Taking expectation under $\pi_{\mathrm{behav}}$ gives
\[
\chi^2(\pi_{\mathrm{new}}\parallel \pi_{\mathrm{behav}})
= \mathbb{E}_{\pi_{\mathrm{behav}}}[(r-1)^2]
\le R^2 \,\mathbb{E}_{\pi_{\mathrm{behav}}}[(\log r)^2]
= R^2 M_2.
\]
\end{proof}

\section{Additional Experiments}
\label{app:exp}

\begin{wrapfigure}{r}{0.4\textwidth}
  \centering
  \includegraphics[width=0.39\textwidth]{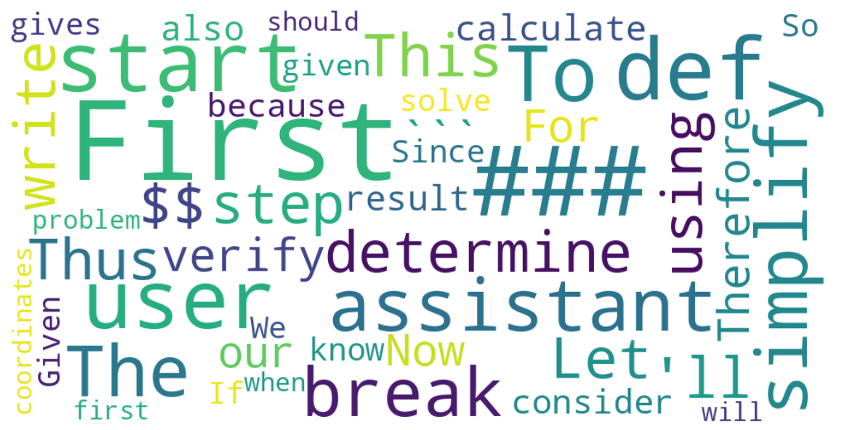}
  \caption{Word clouds of frequently clipped tokens with $\epsilon$ clipping.}
  \label{fig:clipped-tokens} 
\end{wrapfigure}
\textbf{Commonly clipped tokens in GRPO.}
Figure~\ref{fig:clipped-tokens} shows the specific tokens that are most frequently clipped by $\epsilon$-clipping. 
The word cloud of commonly clipped tokens is highly aligned with high-entropy tokens shown in \citep{wang2025beyond}: these tokens are not random or unimportant, but rather belong to the most semantically and structurally critical elements in reasoning traces. Many of them (e.g., First, simplify, determine, To def, Thus, verify, break) are precisely the high-entropy “pivotal tokens” that initiate, connect, or conclude key reasoning steps. Others (e.g., assistant, user, code markers like \#\#\# or \$\$) serve as structural anchors in the dialogue or mathematical formatting.
This observation aligns with Figure~\ref{fig:entropy-ratio}: as the importance weight ratio $|r-1|$ grows, clipped tokens tend to exhibit higher entropy.

\clearpage
\begin{wrapfigure}{r}{0.4\textwidth}
  \centering
  \includegraphics[width=0.39\textwidth]{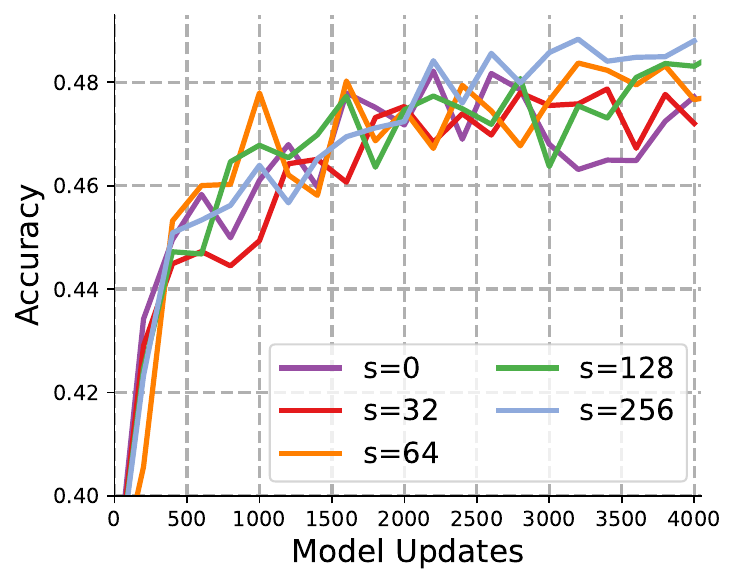}
  \vspace{-0.1cm}
  \caption{The performance of M2PO under different stalenesson Qwen2.5-Math-7B.}
  \label{fig:m2po-staleness} 
  \vspace{-0.3cm}
\end{wrapfigure}
\textbf{M2PO under different staleness.}
In Table~\ref{tab:performance-compare}, we compare the performance between M2PO and other baselines under $s=256$. 
To further study the robustness of M2PO across different staleness levels, we train Qwen2.5-Math-7B with M2PO under $s=0,32,64,128,256$, as shown in Figure~\ref{fig:m2po-staleness}. 
We observe that larger staleness slightly slows down the initial accuracy gain during the very early phase of training, reflecting the expected delay when using outdated rollouts. 
However, this effect quickly diminishes: all curves steadily improve and converge to nearly the same accuracy. Notably, even under extreme staleness ($s=256$), M2PO achieves performance on par with the on-policy case ($s=0$), without signs of collapse or degradation. These results highlight that M2PO effectively preserves the learning signal contained in stale data, enabling stable training and strong generalization across a wide range of staleness values.

\end{document}